\documentclass{article}
\usepackage{spconf,amsmath,graphicx}
\usepackage{multirow}
\usepackage{colortbl}
\usepackage{url}
\usepackage{amssymb}
\makeatletter
\newcommand{\linebreakand}{
  \end{@IEEEauthorhalign}
  \hfill\mbox{}\par
  \mbox{}\hfill\begin{@IEEEauthorhalign}
}
\makeatother

\usepackage{enumitem}
\setlist{nosep, leftmargin=14pt}
\title{Topology and Intersection-Union Constrained Loss Function for Multi-Region Anatomical Segmentation in Ocular Images}

\name{Ruiyu Xia$^{\dagger}$ \quad Jianqiang Li$^{\dagger}$ \quad Xi Xu$^{\dagger *}$ \quad Guanghui Fu$^{\ddagger *}$}

\address{
  $^{\dagger}$ College of Computer Science, Beijing University of Technology, Beijing, China \\
  $^{\ddagger}$ Sorbonne Université, Institut du Cerveau - Paris Brain Institute - ICM, CNRS, Inria, Inserm, AP-HP, \\Hôpital de la Pitié Salpêtrière, F-75013, Paris, France
}

\begin{document}
\maketitle
\begin{abstract}
Ocular Myasthenia Gravis (OMG) is a rare and challenging disease to detect in its early stages, but symptoms often first appear in the eye muscles, such as drooping eyelids and double vision. Ocular images can be used for early diagnosis by segmenting different regions, such as the sclera, iris, and pupil, which allows for the calculation of area ratios to support accurate medical assessments. However, no publicly available dataset and tools currently exist for this purpose. To address this, we propose a new topology and intersection-union constrained loss function (TIU loss) that improves performance using small training datasets. We conducted experiments on a public dataset consisting of 55 subjects and 2,197 images. Our proposed method outperformed two widely used loss functions across three deep learning networks, achieving a mean Dice score of 83.12\% [82.47\%, 83.81\%] with a 95\% bootstrap confidence interval. In a low-percentage training scenario (10\% of the training data), our approach showed an 8.32\% improvement in Dice score compared to the baseline. Additionally, we evaluated the method in a clinical setting with 47 subjects and 501 images, achieving a Dice score of 64.44\% [63.22\%, 65.62\%]. We did observe some bias when applying the model in clinical settings. These results demonstrate that the proposed method is accurate, and our code along with the trained model is publicly available.
\end{abstract}
\section{Introduction} \label{sec:intro}
Myasthenia gravis (MG) is a chronic rare disease caused by neuromuscular transmission disorder~\cite{dresser2021myasthenia}. With timely symptom recognition and treatment, patients can significantly improve their quality of life. MG has five types: ocular, mild systemic, moderate systemic, acute severe, delayed severe, and muscular atrophy~\cite{keesey2004clinical}. Ocular Myasthenia Gravis (OMG) is the mildest form, involving only eye muscle weakness~\cite{nair2014ocular}, with symptoms such as eyelid drooping or diplopia~\cite{ahmed2021role}. An automated method to identify ocular regions can help quantify disease severity, aiding in diagnosis and large-scale analysis. 
Some related research, such as Aayush et al.\cite{chaudhary2019ritnet} and Rot et al.\cite{rot2018deep}, has proposed models for eye structure segmentation in ocular image; however, these works primarily focus on healthy subjects. Currently, no open-source tool exists for this purpose.

Deep learning-based segmentation methods can achieve high performance by learning powerful feature representations and optimizing pixel-level accuracy~\cite{minaee2021image}. However, these methods often face limited training data due to high data collection costs~\cite{tajbakhsh2020embracing} and may overlook biological structures, causing inconsistencies in segmentation results~\cite{hu2019topology}. Topology-based loss functions have been proposed to preserve relationships between regions in medical images~\cite{hu2021topology, fu2024projected}. Fu et al.~\cite{fu2024projected} also showed improved performance using topological loss with small training sets. Despite this, few methods consider the intersections and relationships between regions, which is crucial for segmenting eye regions where the iris surrounds the pupil, and both the pupil and sclera are mutually exclusive.

We propose a new loss function that leverages topology constraints and the intersection and union relationships between eye regions to enhance iris segmentation performance, particularly with small training sets. The method integrates with pixel-level loss functions like cross-entropy or Dice loss and was evaluated using three deep learning models. Developed on a public dataset with healthy subjects, it was tested on both public and clinical datasets. Results consistently showed performance improvements across different training set sizes.

\section{Methods} \label{sec:methods}
\begin{figure}[!hbtp]
\centering
\includegraphics[width=0.9\linewidth]{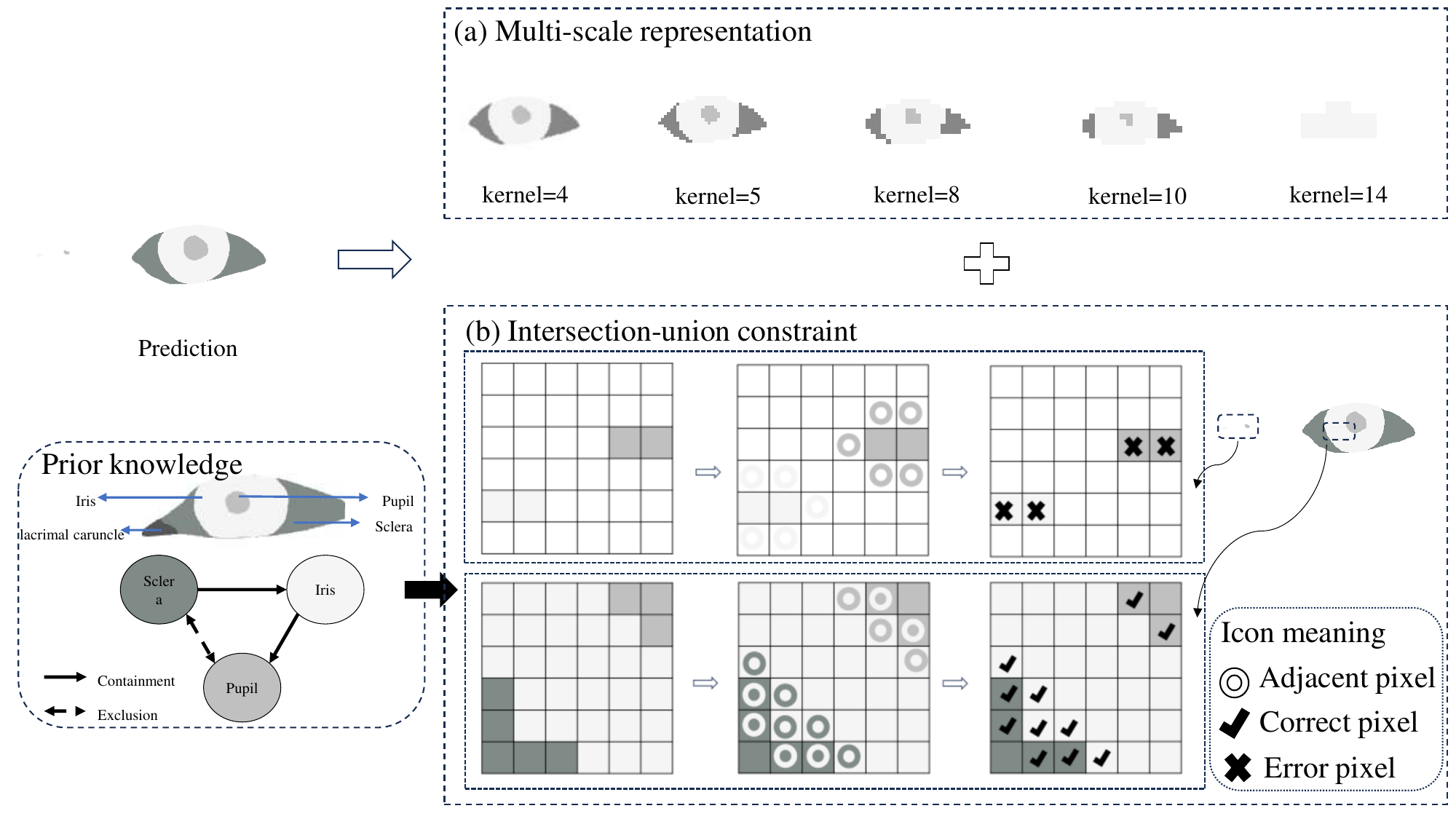}
\caption{Processing flow of the proposed method. This diagram describes the process that the pixels in the predicted image are subject to topological constraints.}
\label{fig:overall_flow}
\end{figure}

The proposed loss function involves two key operations: using MaxPooling for multi-scale pixel representation, and applying ReLU to enforce intersection-union constraints, as shown in Figure~\ref{fig:overall_flow}~(a) and~(b).

\subsection{Multi-scale topology constraint by MaxPooling}\label{sec:methods:comp_extract}
Multi-scale representation has been demonstrated as an effective way to incorporate constraints from multiple perspectives as shown by the topology loss proposed by Fu et al.~\cite{fu2024projected}.
This simplified representation can be efficiently obtained using MaxPooling with different kernel size as illustrated in Figure~\ref{fig:overall_flow}~(a).
The multi-scale representation can effectively enhance the robustness of the model by preserving critical features across various scales. Also, this simple, parameter-free operation makes it computationally efficient and easy to train within the network.

We first normalize the predictions to the range [0, 1], then apply 2D MaxPooling $n$ times with different kernel sizes $K=(k_1, k_2, ...,k_n)$ to both the prediction $P$ and the ground truth $G$. The stride is set to be equal to the respective kernel size to ensure non-overlapping regions. The resulting multi-scale representations for the prediction $P_K$ and the ground truth $G_K$ are defined as follows:
\begin{equation}
\begin{aligned}
    P_K =&\text{MP}^k(P), K=(k_1, k_2, ...,k_n) \\
    G_K =&\text{MP}^k(G), K=(k_1, k_2, ...,k_n)
\end{aligned}
\label{eq:maxpool}
\end{equation}
where $\text{MP}^k$ denotes the MaxPooling operation with kernel size $k$.
We can minimize the difference between the ground truth and predictions at different scales as follows:
\begin{equation}
    L_1=\frac{|P_K-G_K|}{K}
\end{equation}

\subsection{Intersection-union constraint by ReLU} \label{sec:methods:constraint} 
For the predictions $P$ and ground truth $G$, we can supervise the model by incorporating the anatomical relationships between the predicted eye structures: sclera ($P^s$), iris ($P^i$), and pupil ($P^p$), where $P = (P^s, P^i, P^p)$. 
Specifically, the sclera ($P^s$) encloses the iris ($P^i$), the iris ($P^i$) encloses the pupil ($P^p$), and the sclera ($P^s$) and pupil ($P^p$) are mutually exclusive.

For the exclusive regions between $P^s$ and $P^p$, meaning these two regions should not overlap, we introduce a constraint using ReLU as follows:
\begin{equation}
    R_o = \text{ReLU}(\min(P^s, P^p)) = \max(0, \min(P^s, P^p))
\end{equation}
where $R_o$ represents the overlapping regions that need to be minimized.

For the enclosed regions, where $P^s \geq P^i$ and $P^i \geq P^p$, the violation regions, indicating places where these constraints are not satisfied, can be expressed as:
\begin{equation}
\begin{aligned}
    R_v^{s,i} &= \text{ReLU}(P^i - P^s) = \max(0, P^i - P^s), \\
    R_v^{i,p} &= \text{ReLU}(P^p - P^i) = \max(0, P^p - P^i),
\end{aligned}
\end{equation}
where $R_v^{s,i}$ and $R_v^{i,p}$ denote the violation regions that need to be minimized for correct anatomical relationships.
The loss used to enforce the intersection-union constraint is defined as:
% \begin{equation}
%     L_2 = \sum_{k=1}^K(||R_o||_1+||R_v^{s,i}||_1+||R_v^{i,p}||_1)
% \end{equation}
\begin{equation}
    L_2 = ||R_o||_1+||R_v^{s,i}||_1+||R_v^{i,p}||_1
\end{equation}
where $||\cdot||_1$ represents the L1 norm, which sums up all pixel-wise violations.

\subsection{Final loss}

The final loss function can be summarized in three parts by combining the pixel-level loss function $L_p$, such as Dice loss, with the two losses described above:
\begin{equation}
    L_f = \alpha_1 L_1+\alpha_2 L_2+ \beta L_p
\end{equation}
where $\alpha_1$ and $\alpha_2$ as the weights for multi-scale supervision and relationship supervision and $L_p$ is the weight for pixel-level loss.

\section{Experiments and Results} \label{sec:experiments}

\subsection{Dataset} \label{sec:experiments_dataset}
The proposed model was developed by using public dataset, and was evaluated in both public and clinical dataset.
Since no public datasets met the specific requirements for OMG diagnosis, we used the UBIRIS.v2 dataset~\cite{UBIRIS_proencca2009ubiris}\footnote{\url{http://iris.di.ubi.pt/}}, consisting of 11,102 images from 261 subjects, and re-annotated four regions: sclera, iris, pupil, and lacrimal caruncle.
Most subjects were Latin Caucasian (90\%), with smaller proportions of Black (8\%) and Asian (2\%) participants, differing from our target population of Chinese patients. Additionally, the dataset's natural light conditions were inconsistent with our clinical setting. Thus, we selected a subset of 2,196 images from 56 subjects that better suited our needs. The data was split into training (1,759 images, 46 subjects) and test sets (437 images, 10 subjects) with an 8:2 ratio. Images were resized to 480$\times$320 during pre-processing.

\begin{figure}
\centering
\includegraphics[width=1\linewidth]{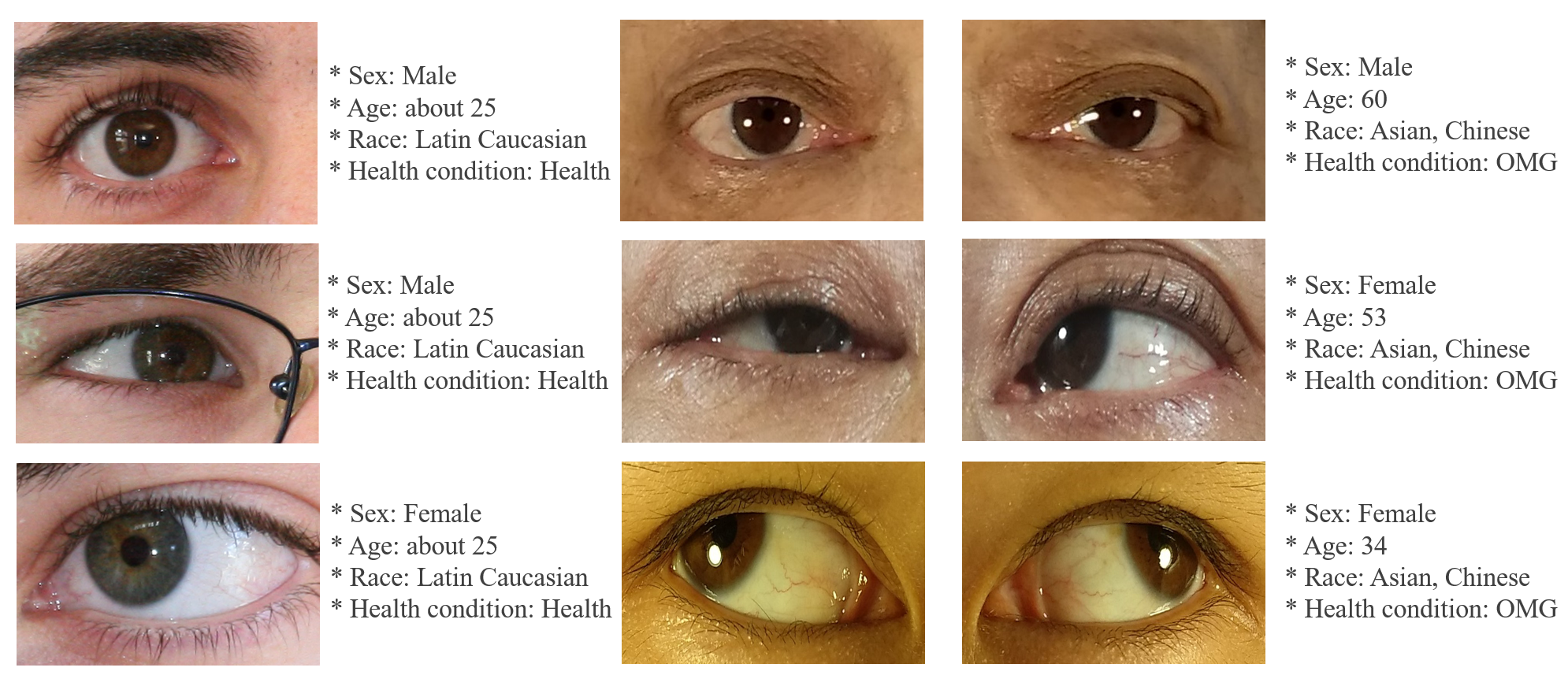}
\caption{Examples of the ocular images from the public dataset (UBIRIS.v2) for healthy subjects (Latin or Caucasian) and the clinical private dataset for subjects diagnosed with OMG disease (Chinese, Asian). }
\label{fig:data_example}
\end{figure}

A prospective study was conducted on 47 myasthenia gravis patients from Beijing Hospital, China. 
The study was approved by the Ethics Committee and relevant authorities (No. 2023BJYYEC-226-01), with the informed consent form signed before filming. The private dataset contains 501 images from 47 patients (23\% male, 77\% female, mean age: 64$\pm$10) collected from January to July 2024. 
These patients, all Chinese, were selected for having ptosis and a clinical diagnosis of ocular myasthenia gravis. Some had only one abnormal eye, resulting in 73 normal eye images and 428 abnormal eye images.
All images were captured with the camera at one meter from the patient's face. The initial data consisted of video recordings, where patients moved their eyes for different angles. Key frames were extracted to form the facial image dataset. Using the Dlib algorithm~\cite{dlib_kavitha2022implementing}\footnote{\url{http://dlib.net/}}, eye regions were localized and cropped to create a clinical dataset. For pre-processing, clinical images were resized to 480$\times$320 to match the public dataset. 
Ground truth labels for both datasets were created using LabelMe~\footnote{\url{https://www.labelme.io}} by five annotators.
Figure~\ref{fig:data_example} shows examples from both datasets, and Table~\ref{tab:dataset} provides the data distribution.

\begin{table}
\centering
\caption{Data distribution of the experimental dataset. Our method was trained using the public dataset (UBIRIS.v2), and evaluated on both public and private dataset. $N_s$ represents the number of subjects, and $N_i$ represents as the number of images. }
\label{tab:dataset}
\begin{tabular}{|l|l|l|l|} 
\hline
Dataset                    & Train/validation or test & $N_s$ & $N_i$  \\ 
\hline
\multirow{2}{*}{UBIRIS.v2} & Train/validation         & 46    & 1759   \\ 
\cline{2-4}
                           & Test                     & 10    & 437    \\ 
\hline
Clinical data              & Test                     & 47    & 501    \\
\hline
\end{tabular}
\end{table}

\subsection{Implementation details}

We evaluate our loss function on three segmentation baselines: UNet~\cite{ronneberger2015u}, FCN~\cite{long2015fully}, and UNTER~\cite{hatamizadeh2022unetr}. UNet is reliable for medical image tasks, FCN provides a foundational approach to pixel-wise prediction, and UNTER uses transformers to enhance segmentation performance.

The proposed loss function can be used as an additional term with any loss function. We used two baselines: cross-entropy loss~\cite{zhang2018generalized}, which is effective for multi-class problems with class imbalance, and Dice loss~\cite{sudre2017generalised}, which optimizes overlap between predicted and actual regions in segmentation tasks.

To assess model performance with limited data and the consistency of our loss function's benefits, we trained using different loss functions on 10\%, 50\%, and 100\% of the training set.
Since the clinical data includes both healthy and OMG-affected eyes, we used the best-performing model from the public dataset to evaluate performance separately for the healthy and OMG groups.

Training and evaluation were conducted on an NVIDIA 24GB RTX 4090 GPU. All models were trained for 100 epochs with a batch size of 16, using early stopping to prevent overfitting in small training sets. We used Dice coefficient and HD95 as metrics, reporting both the mean and 95\% confidence interval (CI), calculated via bootstrap methods on the test set. The codes and trained model are publicly available via: \url{https://github.com/staryoyo/TIU_Loss}

\subsection{Results}
\begin{figure}
    \centering
    \includegraphics[width=1\linewidth]{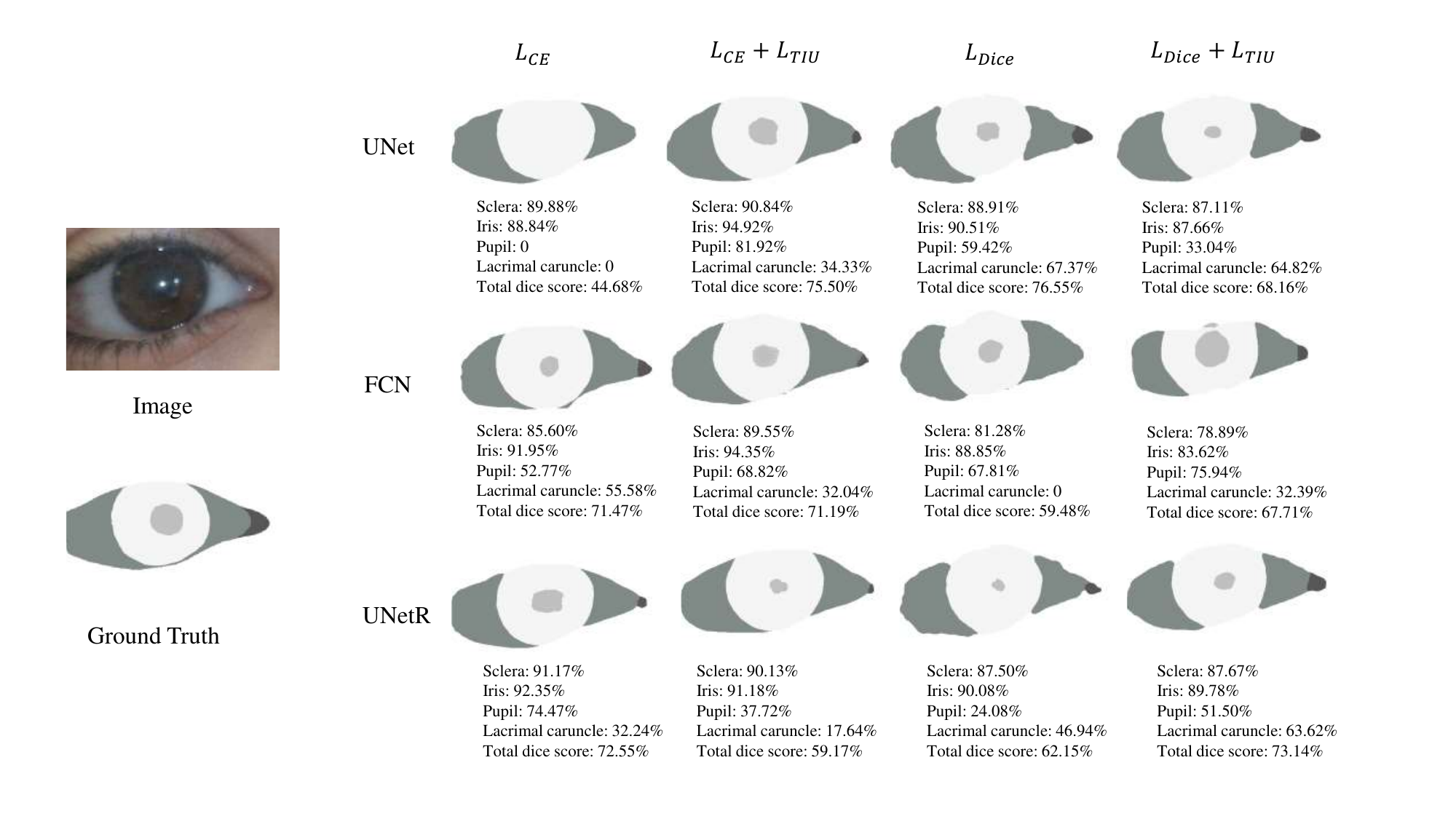}
    \caption{Qualitative results of different models trained with 100\% of training data, compared with different loss combinations on UBIRIS.v2. }
    \label{fig:result:public}
\end{figure}

\begin{figure}
    \centering
    \includegraphics[width=1\linewidth]{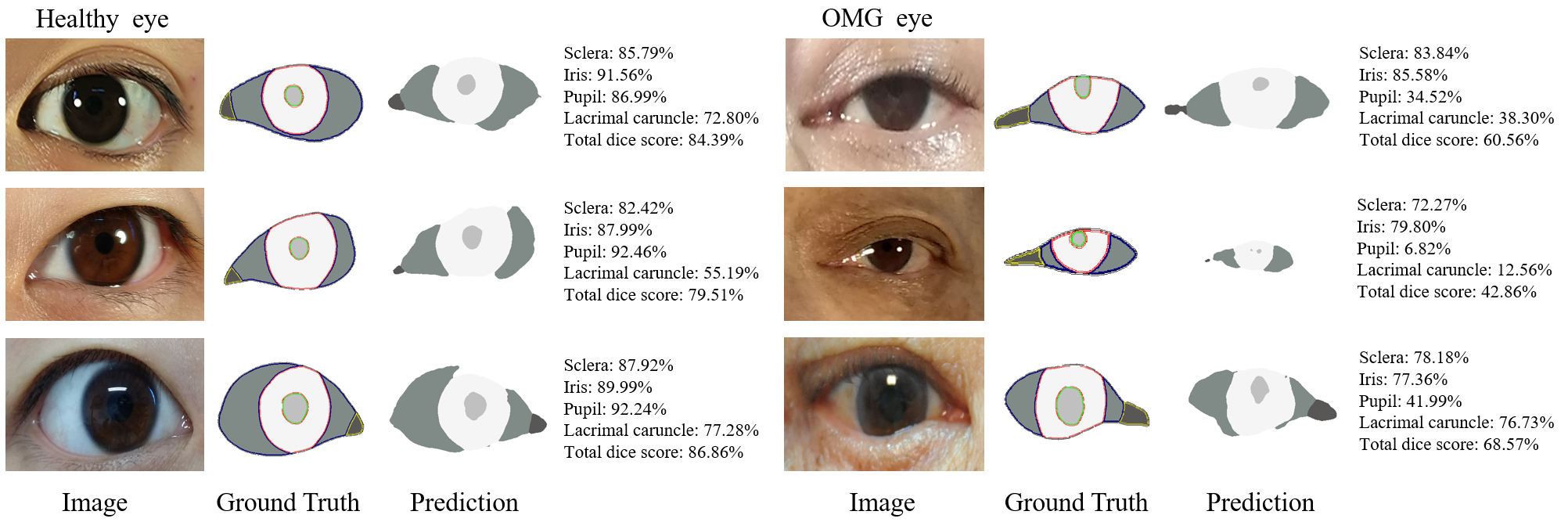}
    \caption{Qualitative results of UNet trained with 100\% of the UBIRIS.v2 dataset using Dice+TIU on Clinical data of the OMG and healthy eye. }
    \label{fig:result:clinical}
\end{figure}

The model trained under different percentage of training set and evaluate on the test set from public and clinical data can be seen in Table~\ref{tab:exp_combine}. The performance of the best model on the public test set, evaluated separately for the healthy and OMG groups, is shown in Table~\ref{tab:contrast}. Examples of the model's performance on the public dataset and clinical dataset are shown in Figure~\ref{fig:result:public} and Figure~\ref{fig:result:clinical}, respectively.

\begin{table}[!ht]
\centering
\caption{The model was trained on different percentages of the training set (10\%, 50\%, and 100\%) from the public dataset (UBIRIS.V2) and evaluated on the test sets from both the public and clinical datasets. $N_i$ represents the number of images in the training set. We report the mean value of Dice score (in \%) and HD95 with a 95\% bootstrap confidence interval.
}
\label{tab:public}
\resizebox{1.0\linewidth}{!}{
\begin{tabular}{|c|c|c|c|c|c|c|} 
\hline
\multicolumn{1}{|c|}{\multirow{2}{*}{Train\% ($N_i$)}} & \multicolumn{1}{c|}{\multirow{2}{*}{Model}} & \multicolumn{1}{c|}{\multirow{2}{*}{Loss}} & \multicolumn{2}{c}{Public data }               & \multicolumn{2}{c}{Clinical data}               \\ 
\cline{4-7}
\multicolumn{1}{|l|}{}                                      & \multicolumn{1}{l|}{}                       & \multicolumn{1}{l|}{}                      & Dice                 & HD95                    & Dice                 & HD95                     \\ 
\hline
\multirow{12}{*}{10\% (220)}                                & \multirow{4}{*}{UNet}                       & CE                                         & 62.72 [61.42, 63.88] & 45.78 [43.15, 48.35]    & 38.35 [37.41, 39.30] & 74.37 [72.22, 76.64]     \\ 
\cline{3-7}
                                                            &                                             & CE+TIU                                     & 65.88 [64.60, 67.17] & 37.57 [35.56, 39.71]    & 40.28 [39.09, 41.42] & 65.75 [63.94, 67.76]     \\ 
\cline{3-7}
                                                            &                                             & Dice                                       & 28.68 [28.02, 29.39] & 221.30 [218.73, 223.81] & 21.55 [20.81, 22.30] & 241.86 [239.50, 244.22]  \\ 
\cline{3-7}
                                                            &                                             & Dice+TIU                                   & 29.46 [28.64, 30.34] & 207.29 [203.67, 211.14] & 24.48 [23.84, 25.21] & 222.59 [219.54, 225.51]  \\ 
\cline{2-7}
                                                            & \multirow{4}{*}{FCN}                        & CE                                         & 45.39 [34.80, 45.93] & 55.53 [51.40, 57.03]    & 24.96 [23.95, 25.89] & 80.18 [78.95, 81.48]     \\ 
\cline{3-7}
                                                            &                                             & CE+TIU                                     & 53.71 [52.53, 54.93] & 45.54 [43.65, 47.58]    & 33.79 [32.84, 34.78] & 68.12 [66.43, 68.84]     \\ 
\cline{3-7}
                                                            &                                             & Dice                                       & 1.23 [1.09, 1.39]    & 195.52 [192.02, 198.67] & 2.32 [1.66, 2.07]    & 191.27 [187.93, 194.58]  \\ 
\cline{3-7}
                                                            &                                             & Dice+TIU                                   & 4.36 [4.08, 4.67]    & 211.63 [209.37, 213.72] & 3.20 [2.93, 3.48]    & 221.47 [218.76, 224.34]  \\ 
\cline{2-7}
                                                            & \multirow{4}{*}{UNetR}                      & CE                                         & 63.89 [62.75, 65.01] & 35.63 [33.95, 37.54]    & 37.40 [36.40, 38.49] & 68.11 [66.25, 70.23]     \\ 
\cline{3-7}
                                                            &                                             & CE+TIU                                     & 67.78 [66.57, 69.00] & 35.00 [32.97, 37.18]    & 39.67 [38.63, 40.72] & 67.14 [65.25, 69.21]     \\ 
\cline{3-7}
                                                            &                                             & Dice                                       & 29.47 [28.79, 30.14] & 210.70 [207.96, 213.48] & 20.98 [20.22, 21.69] & 238.36 [235.77, 241.16]  \\ 
\cline{3-7}
                                                            &                                             & Dice+TIU                                   & 26.89 [26.11, 27.72] & 220.80 [217.82, 223.79] & 22.59 [21.88, 23.24] & 228.68 [225.68, 231.73]  \\ 
\hline
\multirow{12}{*}{50\% (1100)}                               & \multirow{4}{*}{UNet}                       & CE                                         & 80.01 [79.10, 80.84] & 28.27 [26.09, 31.03]    & 57.54 [56.35, 58.73] & 49.02 [47.02, 51.06]     \\ 
\cline{3-7}
                                                            &                                             & CE+TIU                                     & 82.19 [81.27, 83.00] & 24.70 [22.65, 26.92]    & 63.72 [62.62, 64.87] & 40.73 [38.74, 42.91]     \\ 
\cline{3-7}
                                                            &                                             & Dice                                       & 58.89 [58.22, 59.55] & 56.48 [53.95, 58.78]    & 42.17 [41.09, 43.22] & 70.53 [68.47, 72.59]     \\ 
\cline{3-7}
                                                            &                                             & Dice+TIU                                   & 59.32 [58.28, 60.24] & 95.23 [91.38, 99.38]    & 46.20 [45.33, 47.09] & 104.28 [101.29, 107.11]  \\ 
\cline{2-7}
                                                            & \multirow{4}{*}{FCN}                        & CE                                         & 42.60 [42.31, 42.84] & 68.72 [67.59, 70.02]    & 34.20 [33.44, 34.84] & 74.45 [73.29, 75.59]     \\ 
\cline{3-7}
                                                            &                                             & CE+TIU                                     & 43.59 [43.10, 44.05] & 67.12 [65.57, 68.78]    & 34.00 [33.24, 34.87] & 73.73 [72.44, 75.07]     \\ 
\cline{3-7}
                                                            &                                             & Dice                                       & 48.39 [47.53, 49.30] & 59.91 [58.34, 61.53]    & 23.86 [22.84, 24.84] & 83.53 [82.15, 84.98]     \\ 
\cline{3-7}
                                                            &                                             & Dice+TIU                                   & 50.00 [49.15, 50.82] & 61.98 [60.54, 63.98]    & 26.56 [25.57, 27.56] & 82.01 [80.43, 83.48]     \\ 
\cline{2-7}
                                                            & \multirow{4}{*}{UNetR}                      & CE                                         & 81.60 [80.80, 82.44] & 24.57 [22.45, 26.73]    & 60.75 [59.44, 61.95] & 47.39 [45.54, 49.94]     \\ 
\cline{3-7}
                                                            &                                             & CE+TIU                                     & 79.98 [79.01, 80.91] & 22.18 [20.50, 24.09]    & 62.18 [61.01, 63.39] & 39.34 [37.47, 41.05]     \\ 
\cline{3-7}
                                                            &                                             & Dice                                       & 59.37 [58.66, 60.05] & 58.25 [55.58, 60.99]    & 41.72 [40.84, 42.67] & 70.40 [68.46, 72.41]     \\ 
\cline{3-7}
                                                            &                                             & Dice+TIU                                   & 47.80 [47.38, 48.24] & 119.08 [115.29, 122.68] & 42.58 [40.84, 42.67] & 124.57 [121.59, 127.76]  \\ 
\hline
\multirow{12}{*}{100\% (1759)}                              & \multirow{4}{*}{UNet}                       & CE                                         & 80.72 [79.81, 81.51] & 21.43 [19.95, 23.01]    & 58.99 [57.53, 60.33] & 46.43 [44.29, 48.41]     \\ 
\cline{3-7}
                                                            &                                             & CE+TIU                                     & 83.12 [82.47, 83.81] & 23.00 [21.20, 25.11]    & 64.44 [63.22, 65.62] & 40.09 [38.23, 42.02]     \\ 
\cline{3-7}
                                                            &                                             & Dice                                       & 82.66 [51.85, 83.40] & 26.15 [24.11, 28.45]    & 56.19 [54.62, 57.75] & 48.50 [46.57, 50.36]     \\ 
\cline{3-7}
                                                            &                                             & Dice+TIU                                   & 82.43 [81.65, 83.17] & 23.48 [21.81, 25.22]    & 58.48 [57.10, 59.76] & 48.04 [45.97, 50.15]     \\ 
\cline{2-7}
                                                            & \multirow{4}{*}{FCN}                        & CE                                         & 81.84 [80.93, 82.74] & 21.42 [19.80, 23.09]    & 56.79 [55.52, 58.03] & 45.44 [43.51, 47.28]     \\ 
\cline{3-7}
                                                            &                                             & CE+TIU                                     & 80.46 [79.54, 81.35] & 23.44 [21.43, 25.62]    & 58.51 [57.21, 59.81] & 43.95 [41.98, 46.00]     \\ 
\cline{3-7}
                                                            &                                             & Dice                                       & 54.26 [53.36, 55.14] & 50.57 [49.12, 51.95]    & 33.96 [32.65, 35.38] & 69.61 [68.09, 71.31]     \\ 
\cline{3-7}
                                                            &                                             & Dice+TIU                                   & 60.53 [59.25, 61.82] & 44.98 [43.29, 46.81]    & 25.47 [24.00, 26.79] & 81.85 [79.98, 83.77]     \\ 
\cline{2-7}
                                                            & \multirow{4}{*}{UNetR}                      & CE                                         & 74.80 [73.68, 75.80] & 25.95 [24.05, 27.94]    & 53.46 [52.24, 54.72] & 51.53 [49.41, 53.74]     \\ 
\cline{3-7}
                                                            &                                             & CE+TIU                                     & 78.56 [77.60, 79.52] & 27.88 [25.59, 30.17]    & 53.07 [51.74, 54.41] & 50.30 [48.36, 52.40]     \\ 
\cline{3-7}
                                                            &                                             & Dice                                       & 81.73 [80.92, 82.48] & 24.76 [22.98, 26.77]    & 57.59 [56.10, 58.96] & 46.99 [45.11, 49.01]     \\ 
\cline{3-7}
                                                            &                                             & Dice+TIU                                   & 81.27 [80.92, 82.02] & 28.16 [25.88, 30.46]    & 59.60 [58.10, 61.11] & 47.21 [44.98, 49.53]     \\
\hline
\end{tabular}
}
\label{tab:exp_combine}
\end{table}
\begin{table}[!ht]
    \centering
    \caption{Comparative experimental results of diseased eyes and healthy eyes in clinical dataset. The UNet we applied here is the one that achieved the best performance on the clinical dataset: trained using a combination of CE+TIU loss on 100\% of the training data from the open-source dataset.}
    \label{tab:contrast}
    \resizebox{1.0\linewidth}{!}{
    \begin{tabular}{|c|c|c|c|c|c|c|}
    \hline
        \multicolumn{2}{|c|}{} & \multicolumn{2}{|c|}{\textbf{OMG (428)}} & \multicolumn{2}{|c|}{\textbf{Health (73)}} \\ \hline
        \textbf{Model} & \textbf{Loss} & \textbf{Dice  [95\%CI]} & \textbf{HD95\% [95\%CI]} & \textbf{Dice [95\%CI]} & \textbf{HD95\% [95\%CI]}  \\ \hline
        \textbf{\multirow{4}*{Unet}} & CE & 58.47 [57.03, 59.91] & 47.37 [45.03, 49.78] & 62.06 [57.42, 66.41] & 40.92 [36.15, 46.10] \\ \cline{2-6}
        \textbf{} & CE+TIU & 64.33 [63.13, 65.54] & 40.44 [38.37, 42.34] & 65.11 [60.17, 69.90] & 38.07 [33.66, 43.04]   \\ \cline{2-6}
        \textbf{} & Dice & 55.65 [54.17, 57.18] & 48.89 [46.70, 50.94] & 59.38 [53.07, 65.54] & 46.22 [40.59, 52.61]   \\ \cline{2-6}
        \textbf{} & Dice+TIU & 57.56 [56.27, 58.78] & 49.04 [46.67, 51.52] & 63.86 [60.02, 67.78] & 42.22 [36.62, 47.83]  \\ \hline
    \end{tabular}
}
\end{table}

Experimental results show that adding the proposed loss function improves performance across most models. For instance, UNet trained on 10\% of the dataset saw its Dice score increase from 62.72 (CE loss) to 65.88 (CE+TIU), while HD95 decreased from 45.78 to 37.57. Similar trends were seen for FCN and UNetR, confirming the effectiveness of the proposed loss. As expected, performance improved with larger training sets; for UNet with CE+TIU loss, Dice increased from 65.88 (10\%) to 83.12 (100\%), while HD95 decreased from 37.57 to 23.00. Across all models, UNet consistently achieved the best performance, with a Dice score of 83.12 at 100\%, outperforming FCN (80.46) and UNetR (78.56), suggesting that UNet is better at capturing complex features, especially when enhanced with the proposed loss function.

Across all models and training set sizes, performance on the clinical dataset was lower than on UBIRIS.v2, highlighting the challenge of domain shift. For example, UNet trained with 100\% of UBIRIS.v2 using CE loss achieved a Dice score of 80.72 on UBIRIS.v2 but only 58.99 on the clinical dataset. The proposed TIU loss effectively improved clinical dataset performance; UNet trained on 10\% of data improved its Dice score from 38.35 (CE loss) to 40.28 (CE+TIU), and HD95 decreased from 74.37 to 65.75. Larger training set portions led to better clinical dataset performance; for instance, UNet trained with 50\% data achieved a Dice score of 63.72, compared to 40.28 with only 10\%. Despite improvements, a performance gap persists due to domain shift.

To investigate clinical dataset performance differences, we divided the data into OMG and healthy eyes. Table~\ref{tab:contrast} shows results for the UNet model trained with 100\% of UBIRIS.v2, which performed best on the clinical dataset. The model performed better on healthy eyes than on OMG, with a Dice score of 65.11 vs. 64.33, and lower HD95 (38.07 vs. 40.44), indicating easier segmentation for healthy eyes. The TIU loss improved performance for both groups: OMG Dice improved from 58.47 (CE) to 64.33 (CE+TIU), with HD95 reduced from 47.37 to 40.44. For healthy eyes, Dice increased from 62.06 to 65.11, and HD95 decreased from 40.92 to 38.07, confirming TIU's benefits in both challenging and easier scenarios.

\section{Conclusion} \label{sec:conclusion}
In this paper, we introduce a new loss function that accounts for the relationships between different eye regions and improves performance when combined with a pixel-level loss function. We conducted experiments using different training set sizes and evaluated the models on both public and clinical datasets. While performance decreased when applying the trained model to the clinical dataset, the proposed loss function still achieved the best results. Future work will focus on training a better-performing model specifically for clinical analysis. All code and trained models are publicly available to support further research.

\section{Compliance with ethical standards}
The institutional ethical standard committee approved the study (No. 2023BJYYEC-226-01).  
All participants gave written informed consent.

\section{Acknowledgments}\label{sec:acknowledgments}
This study is supported by the National High Level Hospital Clinical Research Funding(no.BJ-2023-111). The authors have no relevant financial or non-financial interests to disclose.

\bibliographystyle{IEEEbib}
\bibliography{refs}

\begin{thebibliography}{10}

\bibitem{dresser2021myasthenia}
Laura Dresser, Richard Wlodarski, Kourosh Rezania, and Betty Soliven,
\newblock ``Myasthenia gravis: epidemiology, pathophysiology and clinical manifestations,''
\newblock {\em Journal of clinical medicine}, vol. 10, no. 11, pp. 2235, 2021.

\bibitem{keesey2004clinical}
John~C Keesey,
\newblock ``Clinical evaluation and management of myasthenia gravis,''
\newblock {\em Muscle \& Nerve: Official Journal of the American Association of Electrodiagnostic Medicine}, vol. 29, no. 4, pp. 484--505, 2004.

\bibitem{nair2014ocular}
Akshay~Gopinathan Nair and et~al.,
\newblock ``Ocular myasthenia gravis: a review,''
\newblock {\em Indian journal of ophthalmology}, vol. 62, no. 10, pp. 985--991, 2014.

\bibitem{ahmed2021role}
AL-Bulushi Ahmed, Issa Al~Salmi, Fatma Al~Rahbi, AbdulAziz Al~Farsi, and Suad Hannawi,
\newblock ``The role of thymectomy in myasthenia gravis: A programmatic approach to thymectomy and perioperative management of myasthenia gravis,''
\newblock {\em Asian Journal of Surgery}, vol. 44, no. 6, pp. 819--828, 2021.

\bibitem{chaudhary2019ritnet}
Aayush~K Chaudhary, Rakshit Kothari, and e~tal.,
\newblock ``Ritnet: Real-time semantic segmentation of the eye for gaze tracking,''
\newblock in {\em 2019 IEEE/CVF International Conference on Computer Vision Workshop (ICCVW)}. IEEE, 2019, pp. 3698--3702.

\bibitem{rot2018deep}
Peter Rot, {\v{Z}}iga Emer{\v{s}}i{\v{c}}, Vitomir Struc, and Peter Peer,
\newblock ``Deep multi-class eye segmentation for ocular biometrics,''
\newblock in {\em 2018 IEEE international work conference on bioinspired intelligence (IWOBI)}. IEEE, 2018, pp. 1--8.

\bibitem{minaee2021image}
Shervin Minaee, Yuri Boykov, Fatih Porikli, Antonio Plaza, Nasser Kehtarnavaz, and Demetri Terzopoulos,
\newblock ``Image segmentation using deep learning: A survey,''
\newblock {\em IEEE transactions on pattern analysis and machine intelligence}, vol. 44, no. 7, pp. 3523--3542, 2021.

\bibitem{tajbakhsh2020embracing}
Nima Tajbakhsh, Laura Jeyaseelan, Qian Li, Jeffrey~N Chiang, Zhihao Wu, and Xiaowei Ding,
\newblock ``Embracing imperfect datasets: A review of deep learning solutions for medical image segmentation,''
\newblock {\em Medical image analysis}, vol. 63, pp. 101693, 2020.

\bibitem{hu2019topology}
Xiaoling Hu, Fuxin Li, Dimitris Samaras, and Chao Chen,
\newblock ``Topology-preserving deep image segmentation,''
\newblock {\em Advances in neural information processing systems}, vol. 32, 2019.

\bibitem{hu2021topology}
Xiaoling Hu, Yusu Wang, Li~Fuxin, Dimitris Samaras, and Chao Chen,
\newblock ``Topology-aware segmentation using discrete morse theory,''
\newblock {\em arXiv preprint arXiv:2103.09992}, 2021.

\bibitem{fu2024projected}
Guanghui Fu, Rosana El~Jurdi, Lydia Chougar, Didier Dormont, Romain Valabregue, St{\'e}phane Leh{\'e}ricy, Olivier Colliot, and ICEBERG~Study Group,
\newblock ``Projected pooling loss for red nucleus segmentation with soft topology constraints,''
\newblock {\em Journal of Medical Imaging}, vol. 11, no. 4, pp. 044002--044002, 2024.

\bibitem{UBIRIS_proencca2009ubiris}
Hugo Proen{\c{c}}a, Silvio Filipe, Ricardo Santos, Joao Oliveira, and Luis~A Alexandre,
\newblock ``The ubiris. v2: A database of visible wavelength iris images captured on-the-move and at-a-distance,''
\newblock {\em IEEE Transactions on Pattern Analysis and Machine Intelligence}, vol. 32, no. 8, pp. 1529--1535, 2009.

\bibitem{dlib_kavitha2022implementing}
R~Kavitha, P~Subha, R~Srinivasan, and M~Kavitha,
\newblock ``Implementing opencv and dlib open-source library for detection of driver’s fatigue,''
\newblock in {\em Proc.ICIDCA 2021}, pp. 353--367. Springer, 2022.

\bibitem{ronneberger2015u}
Olaf Ronneberger, Philipp Fischer, and Thomas Brox,
\newblock ``U-net: Convolutional networks for biomedical image segmentation,''
\newblock in {\em Proc.MICCAI 2015}. Springer, 2015, pp. 234--241.

\bibitem{long2015fully}
Jonathan Long, Evan Shelhamer, and Trevor Darrell,
\newblock ``Fully convolutional networks for semantic segmentation,''
\newblock in {\em Proc.CVPR}, 2015, pp. 3431--3440.

\bibitem{hatamizadeh2022unetr}
Ali Hatamizadeh, Yucheng Tang, Vishwesh Nath, Dong Yang, Andriy Myronenko, Bennett Landman, Holger~R Roth, and Daguang Xu,
\newblock ``{UNETR}: Transformers for 3d medical image segmentation,''
\newblock in {\em Proc.WACV}, 2022, pp. 574--584.

\bibitem{zhang2018generalized}
Zhilu Zhang and Mert Sabuncu,
\newblock ``Generalized cross entropy loss for training deep neural networks with noisy labels,''
\newblock {\em Advances in neural information processing systems}, vol. 31, 2018.

\bibitem{sudre2017generalised}
Carole~H Sudre, Wenqi Li, Tom Vercauteren, Sebastien Ourselin, and M~Jorge~Cardoso,
\newblock ``Generalised dice overlap as a deep learning loss function for highly unbalanced segmentations,''
\newblock in {\em Deep Learning in Medical Image Analysis and Multimodal Learning for Clinical Decision Support, DLMIA and ML-CDS, MICCAI 2017}. Springer, 2017, pp. 240--248.

\end{thebibliography}
\end{document}